\documentclass[11pt,a4paper]{article}
\usepackage{hyperref}
\usepackage[hyperref]{acl2018}
\usepackage{times}
\usepackage{latexsym}

\usepackage{url}

\usepackage{graphicx}

\usepackage{amsmath}
\usepackage{bm}

\usepackage{amssymb}

\usepackage{cleveref}
\crefname{section}{§}{§§}
\Crefname{section}{Section}{}
\Crefname{figure}{Fig.}{}
\Crefname{algorithm}{Algorithm}{}
\Crefname{equation}{Equation}{}

\usepackage{booktabs}

\usepackage{multirow}

\usepackage{array}
\newcolumntype{L}[1]{>{\raggedright\let\newline\\\arraybackslash\hspace{0pt}}m{#1}}
\newcolumntype{C}[1]{>{\centering\let\newline\\\arraybackslash\hspace{0pt}}m{#1}}
\newcolumntype{R}[1]{>{\raggedleft\let\newline\\\arraybackslash\hspace{0pt}}m{#1}}

\aclfinalcopy 

\title{Cross-lingual Semantic Parsing}

\author{Sheng Zhang\\
  Johns Hopkins University\\
  {\tt zsheng2@jhu.edu} \\\And
  Kevin Duh\\
  Johns Hopkins University\\
  {\tt kevinduh@cs.jhu.edu} \\\And
  Benjamin Van Durme\\
  Johns Hopkins University\\
  {\tt vandurme@cs.jhu.edu} \\}

\date{}

\begin{document}
\maketitle
\begin{abstract}
We introduce the task of cross-lingual semantic parsing: mapping content provided in a source language into a meaning representation based on a target language.  We present:
    (1) a  meaning representation designed to allow systems to target varying levels of structural complexity (shallow to deep analysis),
    (2) an evaluation metric to measure the similarity between system output and reference meaning representations,
    (3) an end-to-end model with a novel copy mechanism that supports intra-sentential coreference,
    and (4) an evaluation dataset where experiments show our model outperforms
    strong baselines by at least 1.18 $F_1$ score.

\end{abstract}

\section{Introduction}
We are concerned here with representing the semantics of multiple natural languages in a single meaning representation.  Renewed interest in meaning representations has led to a surge of proposed new frameworks,
e.g., GMB~\cite{basile2012developing}, AMR~\cite{banarescu2013abstract},
UCCA~\cite{abend-rappoport:2013:ACL2013},
and UDS~\cite{white-EtAl:2016:EMNLP2016}, as well as further calls to attend to existing representations, e.g., Episodic Logic (EL) \cite{Schubert:2000:ELM:342652.342677,schubert2000situations,10.1007/BFb0013992,schubert:2014:W14-24}, or Discourse Representation Theory (DRT) \cite{kamp1981theory,heim:1988-}.

Many of these efforts are limited to the analysis of English, but with a number of exceptions, e.g., recent efforts by \newcite{DBLP:conf/eacl/BosEBAHNLN17}, ongoing efforts in Minimal Recursion Semantics (MRS) \cite{Copestake95translationusing}, multilingual FrameNet annotation and parsing~\cite{Fung04biframenet:bilingual,Pado:2005:CPR:1220575.1220683}, among others.
For many languages, semantic analysis
can not be performed directly, owing to a lack of training data.  While there is active work in the community focused on rapid construction of resources for low resource languages~\cite{STRASSEL16.1138}, it remains an expensive and perhaps infeasible solution to assume in-language annotated resources for developing semantic parsing technologies.
In contrast, bitext is easier to get: it occurs often without researcher involvement,\footnote{For example, owing to a government decree.} and even when not available, it may be easier to find bilingual speakers that can translate a text, than it is to find experts that will create in-language semantic annotations.  In addition, we are simply further along in being able to automatically understand English than we are other languages, resulting from the bias in investment in English-rooted resources. 

\begin{figure}[t]
\centering
\includegraphics[width=0.49\textwidth]{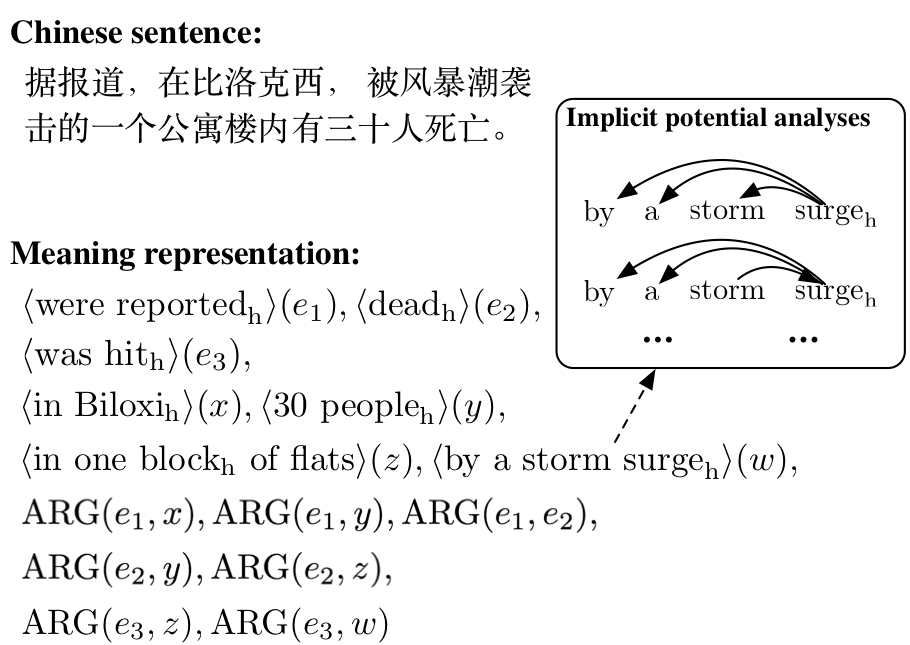}
\caption{Input and output of cross-lingual semantic parsing. 
    The reference translation for the input is ``{\em In Biloxi,
    30 people were reported dead in one block of flats which was hit by a storm surge}''.
    \label{fig:task}}
\end{figure}

Therefore, we propose the task of cross-lingual semantic parsing,
which aims at transducing a sentence in the source language
(e.g., Chinese sentence in \Cref{fig:task}) into
a meaning representation derived from English examples, via bitext.
Our contributions are four-fold:

\noindent(1) We present a meaning representation, which allows systems
to target varying levels of structural complexity (shallow to deep analysis).

\noindent(2) We design an evaluation metric to 
    measure the similarity between system output and reference meaning representations.

\noindent(3) We propose an encoder-decoder model to learn
end-to-end cross-lingual semantic parsing. With a copying mechanism,
    the model is able to solve intra-sentential coreference explicitly.

\noindent(4) We release the first evaluation dataset for cross-lingual semantic parsing.
    Experiments show that our proposed model achieves an $F_1$ score
of 38.38, which outperforms several strong baselines.

\section{Related Work}
Our work synthesizes two strands of research, 
meaning representation and cross-lingual learning.

The meaning representation targeted in this work is akin to that of
\newcite{hobbs2003}, but our eventual goal is to transduce texts from
arbitrary human languages into a \emph{``...broad, language-like,
inference-enabling [semantic representation] in the spirit of
Montague...''}~\cite{AAAI1510051}.  Unlike efforts such as by Schubert and colleagues that directly target such a representation, we are pursuing a strategy that incrementally increases the
complexity of the target representation in accordance with our ability
to fashion models capable of producing it.\footnote{E.g., in \Cref{fig:task} we recognize ``\emph{by a storm surge}'' as an initial structural unit, with multiple potential analysis, which may be further refined based on the capabilities of a given cross-lingual semantic parser.}
Embracing underspecification in the name of tractability is 
exemplified by MRS \cite{copestake2005minimal,copestake:2009:EACL}, the so-called \emph{slacker semantics}, and we draw inspiration from that work.
Representations such as AMR~\cite{banarescu2013abstract} also
make use of underspecification, but usually this is only implicit:
certain aspects of meaning are simply not annotated.  Unlike AMR, but
akin to decisions made in PropBank \cite{palmer2005proposition} (which forms the
majority of the AMR ontological backbone), we target a representation
with a close correspondence to natural language syntax.
Unlike interlingua~\cite{Mitamura91anefficient,And02interlinguaapproximation} that maps the source language into an intermediate representation, and
then maps it into the target language, we
are not concerned with generating text from the meaning representation.
Substantial prior work on meaning representations exists, including
HPSG-based representations~\cite{copestake2005minimal},
CCG-based representations~\cite{Steedman:2000:SP:332037,baldridge-kruijff:2002:ACL,Bos:2004:WSR:1220355.1220535},
and Universal Dependencies based representations~\cite{white-EtAl:2016:EMNLP2016,reddy-EtAl:2017:EMNLP2017}.  See ~\cite{AAAI1510051,abend-rappoport:2017:Long} for further discussion.

Cross-lingual learning has previously been applied to various NLP tasks.
\newcite{Yarowsky:2001:IMT:1072133.1072187,Pado:2009:CAP:1734953.1734960,C16-1056,faruqui-kumar:2015:NAACL-HLT} 
focused on projecting existing annotations on source-language text to the target language.
\newcite{zeman2008cross,ganchev-gillenwater-taskar:2009:ACLIJCNLP,mcdonald-petrov-hall:2011:EMNLP,naseem-barzilay-globerson:2012:ACL2012,Q14-1005} enabled model transfer by sharing
features or model parameters for different languages.
\newcite{sudo2004cross,zhang-duh-vandurme:2017:EACLshort}
worked on cross-lingual information extraction and demonstrated the advantages of
end-to-end learning approaches. In this work, we explore end-to-end learning
for cross-lingual semantic parsing, as discussed in 
\Cref{sec:model}.

\begin{figure*}[t]
\centering
\includegraphics[width=0.99\textwidth]{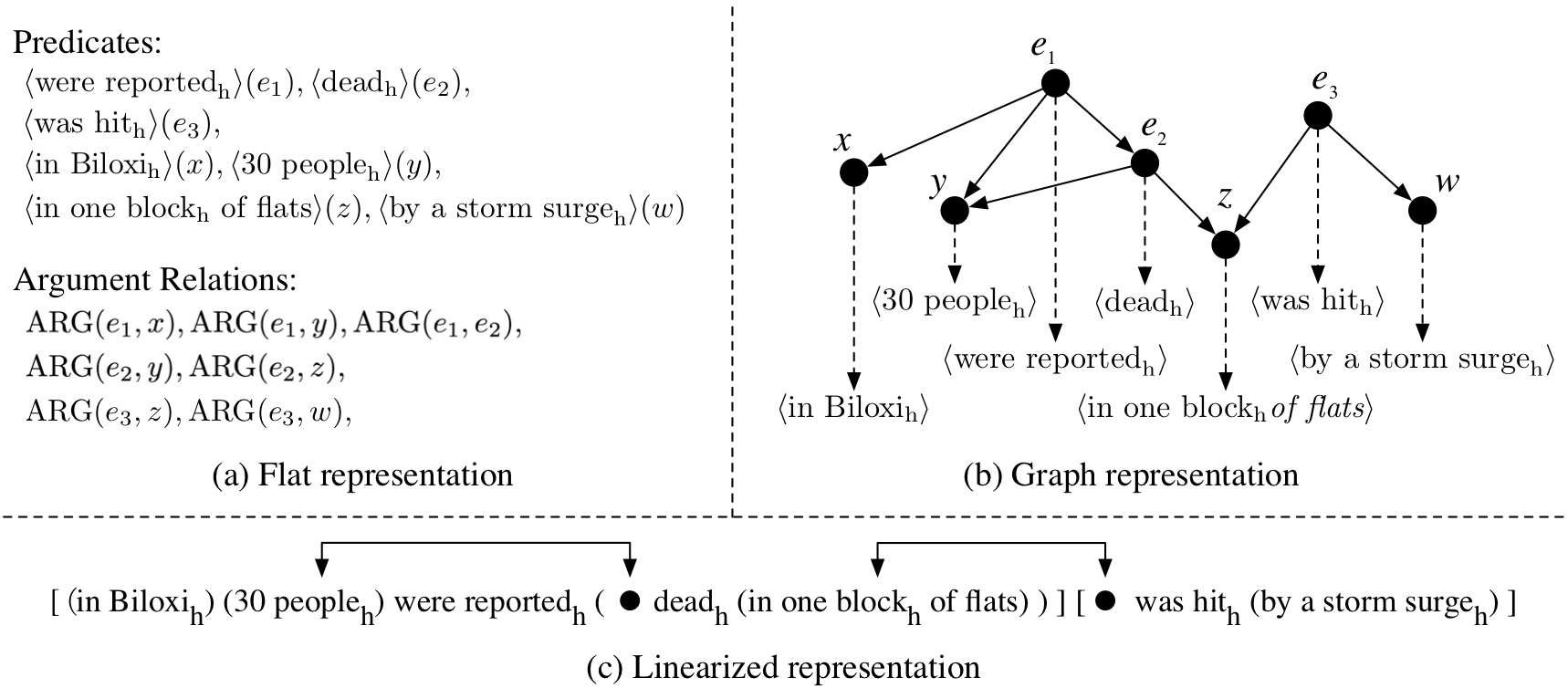}
\caption{Meaning representations.
    \label{fig:repr}}
\end{figure*}

\section{Meaning Representation}
\label{sec:repr}
The goal of cross-lingual semantic parsing is to provide a meaning
representation which can be used for various types of deep and shallow analysis on the target language side.
Many meaning representations potentially suitable for this goal, e.g.,
AMR~\cite{banarescu2013abstract}, UCCA~\cite{abend-rappoport:2013:ACL2013},
UDS~\cite{white-EtAl:2016:EMNLP2016},
and UD\textsc{epLambada}~\cite{reddy-EtAl:2017:EMNLP2017}.
In this work, we choose the  representation used as a scaffold by UDS, namely the PredPatt meaning representation. Other meaning representations may also be feasible.

PredPatt is a framework which defines a set of patterns for shallow semantic
parsing.
The reasons for choosing PredPatt meaning representation are three-fold:
(1) \textbf{Compatibility}: The PredPatt meaning representation relates to Robust Minimal Recursion Semantics (RMRS)~\cite{copestake2007semantic},
aiming for a maximal degree of semantic compatibility.
With such a meaning representation, shallow analysis, such as predicate-argument extraction~\cite{zhang-duh-vandurme:2017:EACLshort}, can be regarded as producing
a semantics which is underspecified and reusable with respect to
deeper analysis, such as lexical semantics and inference~\cite{white-EtAl:2016:EMNLP2016}.
(2) \textbf{Robustness and Speed}: 
Patterns defined in PredPatt for producing
this meaning representation are non-lexical and
linguistically well-founded, and PredPatt has been shown to be fast and accurate enough
to process large volumes of text~\cite{zhang-EtAl:2017:IWCS}.
(3) \textbf{Cross-lingual validity}: Patterns in PredPatt are purely based
on Universal Dependencies, which is designed to
be cross-linguistically consistent. 

In the following sections, we describe three forms of PredPatt meaning
representation (\Cref{fig:repr}). They are created for different purposes,
and are inter-convertible.
In this work, the graph representation is used for evaluation, and the linearized
representation is used for learning cross-lingual semantic parsing.

\subsection{Flat Representation}
The non-recursive or ``flat'' representation can be viewed as a  Parson-style \cite{parsons1990events} and underspecified version of 
neo-Davidsonianized RMRS~\cite{copestake2007semantic}.
As shown in \Cref{fig:repr}(a), the flat representation is a tuple
$\mathcal{F}=\langle P, A\rangle$
where $P$ is a bag of predicates that are all maximally unary,
and $A$ is a bag of arguments represented by separate binary relations.

\noindent{\bf Predicate}: Predicates in PredPatt representation are referred as
{\em complex predicates}: they are open-class predicates represented in the
target language. Scope and lexical information in the predicates are left
unresolved, yet can be recovered incrementally in deep semantic parsing.
From the perspective of RMRS, complex predicates are conjunctions of
underspecified {\em elementary predications}
\cite{copestake2005minimal} where handles are ignored,
but syntax properties from Universal Dependencies are retained.
For instance, in \Cref{fig:repr}(a),
the subscript ``$\text{h}$'' in the predicate ``$\langle$were reported$_\text{h}\rangle$''
indicates that ``reported'' is a syntactic head in the predicate.

\noindent{\bf Argument Relation}: The Parson-style flat representation makes
arguments first-class predications $\textsc{arg}(\cdot, \cdot)$.
Using this style allows incremental addition of arguments,
which is useful in shallow semantics where the arity of open-class predicate
and the argument indexation are underspecified.
They can be recovered when lexicon is available in deep analysis~\cite{dowty1989semantic,copestake2007semantic}.

\subsection{Graph Representation}
\label{sec:graph}
The graph representation as shown in \Cref{fig:repr}(b) is developed to improve
ease of readability, parser evaluation, and integration with lexical semantics.
The structure of the graph representation is a triple $\mathcal{G}=\langle V, I, R\rangle$:
a set of variables $V$ (e.g., $e_1$ and $x$), a mapping $I$ from each variable to its instance
in the target language (e.g., the dotted arrows in \Cref{fig:repr}(b)), and
a mapping $R$ from each pair of variables to their argument relation
(e.g., the solid arrows in \Cref{fig:repr}(b)).
The graph representation can be viewed as an underspecified version of
Dependency Minimal Recursion Semantics (DMRS) \cite{copestake:2009:EACL} due to
the underspecification of scope. Different from DMRS, the graph representation
can be linked cleanly to the syntax of Universal Dependencies in PredPatt.

\subsection{Linearized Representation}
The linearized representation aims to facilitate learning of
cross-lingual semantic parsing. Recently parsers based on recurrent neural
networks that make use of linearized representation have achieved state-of-the-art performance
in constituency parsing~\cite{vinyals2015grammar},
logical form prediction~\cite{dong-lapata:2016:P16-1,jia-liang:2016:P16-1},
cross-lingual open information extraction~\cite{zhang-duh-vandurme:2017:EACLshort},
and AMR parsing~\cite{barzdins-gosko:2016:SemEval,peng-EtAl:2017:EACLlong1}.
An example of PredPatt linearized representation is shown in \Cref{fig:repr}(c): 
Starting at the root node of the dependency tree (i.e., ``reported$_\text{h}$''),
we take an in-order traversal of its spanning tree.
As the tree is expanded, brackets are inserted to denote the beginning or end
of a predicate span, and parentheses are inserted to denote the beginning or end
of an argument span.
The subscript ``$\text{h}$'' indicates the syntactic head of each span.
Intra-sentential coreference occurs when an argument refers to one of its preceding nodes, where
we replace the argument with a special symbol ``$\bullet$'' and add a coreference
link between ``$\bullet$'' and its antecedent.
Such a linearized representation can be viewed as a sequence of tokens accompanied
by a list of conference links.
Brackets, parentheses, syntactic heads, and the special symbol ``$\bullet$''
are all considered as tokens in this representation.

\section{Evaluation Metric $\bm{S}_{(\phi,\psi)}$}
\label{sec:metric}
In cross-lingual semantic parsing,
meaning representation for the target language can be represented in three forms
as shown in \Cref{fig:repr}.
Evaluation of such forms is crucial to the development of algorithms for 
cross-lingual semantic parsing. 
However, there is no method directly available for evaluation.
Related methods come from semantic parsing, whose results are mainly evaluated in three ways:
(1) task correctness~\cite{tang2001using}, which evaluates on a specific NLP task that uses
the parsing results; (2) whole-parse correctness~\cite{Zettlemoyer:2005:LMS:3020336.3020416},
which counts the number of parsing results that are completely correct;
and (3) Smatch~\cite{cai-knight:2013:Short}, which computes the similarity
between two semantic structures.

Nevertheless, in cross-lingual semantic parsing where instances of predicates
are represented in the target language, we need
an evaluate metric that can be used regardless of specific tasks or domains,
and is able to differentiate two parsing results
that have not only similar structures but also similar predicate instances.
We design an evaluation metric $\bm{S}_{(\phi,\psi)}$ that computes the similarity
between two graph representations as shown \Cref{fig:repr}(b).
$\phi$ is the function to score the similarity
between two instances, and $\psi$ scores the similarity
between two argument relations. These scores are normalized to [0, 1]. 

As described in \Cref{sec:graph}, the graph representation consists of three types of
information $\mathcal{G}=(V, I, R)$.
For two graphs $\mathcal{G}_1=(V_1, I_1, R_1)$ and
$\mathcal{G}_2=(V_2, I_2, R_2)$, we define the score $\bm{S}_{(\phi,\psi)}$ to
measure the similarity of $\mathcal{G}_1$ against $\mathcal{G}_2$:
\begin{align*}
    \bm{S}_{(\phi,\psi)}(\mathcal{G}_1, \mathcal{G}_2)\!=\!\max_{m\in \mathcal{M}}\!\big[\!\sum\limits_{v_i \in V_1}\phi\bm{(}I_1(v_i), I_2(m(v_i))\bm{)} \\
     + \sum_{(v_i, v_j)\in U(R_1)}\psi\bm{(}R_1(v_i, v_j), R_2(m(v_i), m(v_j))\bm{)} \Big]
\end{align*}
where $m$ is a mapping from variables in $V_1$ to variables in $V_2$.
$U(R_1)$ is the domain of $R_1$, i.e., all argument edges in $\mathcal{G}_1$. 
$\bm{S}_{(\phi,\psi)}$ computes the highest
similarity score among all possible mappings $\mathcal{M}$.

The precision and recall are computed by
\begin{align*}
    \text{Precision}=\frac{\bm{S}_{(\phi,\psi)}(\mathcal{G}_1, \mathcal{G}_2)}{|U(I_1)| + |U(R_1)|} \\
    \text{Recall}=\frac{\bm{S}_{(\phi,\psi)}(\mathcal{G}_1, \mathcal{G}_2)}{|U(I_2)| + |U(R_2)|}
\end{align*}
where $|U(I_1)|$ is the number of instances in $\mathcal{G}_1$,
$|U(R_1)|$ is the number of argument relations in $\mathcal{G}_1$.

In this work, we set $\phi=\textsc{Bleu}$~\cite{papineni2002bleu} and
$\psi=\delta$, the Kronecker delta. 
\textsc{Bleu} is a widely-used metric in machine translation,
and here it gives partial credits to instance similarity in $\bm{S}_{(\phi,\psi)}$.\footnote{Future work could consider, e.g., a modified {\sc BLEU} that considers Levenshtein distance between tokens for a more robust partial-scoring in the face of transliteration errors.}
Finding an optimal variable mapping $m$ that yields the highest similarity score
$\bm{S}_{(\phi,\psi)}$ is NP-complete.
We instead adopt a strategy used in Smatch~\cite{cai-knight:2013:Short} that
does a hill-climbing search with smart initialization plus 4 random restarts,
and has been shown to give the best trade-off between accuracy and speed.
Smatch for evaluating semantic structures can be
considered as a special case of $\bm{S}_{(\phi,\psi)}$, where $\phi=\delta$
and $\psi=\delta$. We show an example of evaluating two similar graphs using $\bm{S}_{(\phi,\psi)}$
in the supplemental material.

\section{Task}
\label{sec:task}
We formulate the task of cross-lingual semantic parsing as a joint problem 
of sequence-to-sequence learning and coreference resolution. The input is 
a sentence $X$ in the source language, e.g., the Chinese sentence in \Cref{fig:task}.
The output is a linearized meaning
representation as shown in \Cref{fig:repr}(c):
it contains a sequence of tokens $Y$ in the target language as well as
coreference assignments $A$ for each special symbol ``$\bullet$'' in $Y$.

Formally, let the input be a sequence of tokens $X=x_1,\ldots,x_{N}$,
and let the output be a sequence of tokens $Y=y_1,\ldots,y_{M}$ and
a list of coreference assignments $A=[a_1,\ldots,a_M]$,
where $a_t$ is the coreference assignment for $y_t$.
The set of possible assignments for $y_t$
is $\mathcal{A}(t) = \{\epsilon, y_1,\ldots,y_{t-1}\}$, a dummy
antecedent $\epsilon$ and all preceding tokens.
The dummy antecedent $\epsilon$ represents a scenario where the token is not a
special symbol ``$\bullet$'' and should be assigned to none of the preceding
tokens. $N$ is the length of the input sentence, and $M$, the length of
the output sentence.

\section{Model}
\label{sec:model}

\begin{figure*}[t]
\centering
\includegraphics[width=0.99\textwidth]{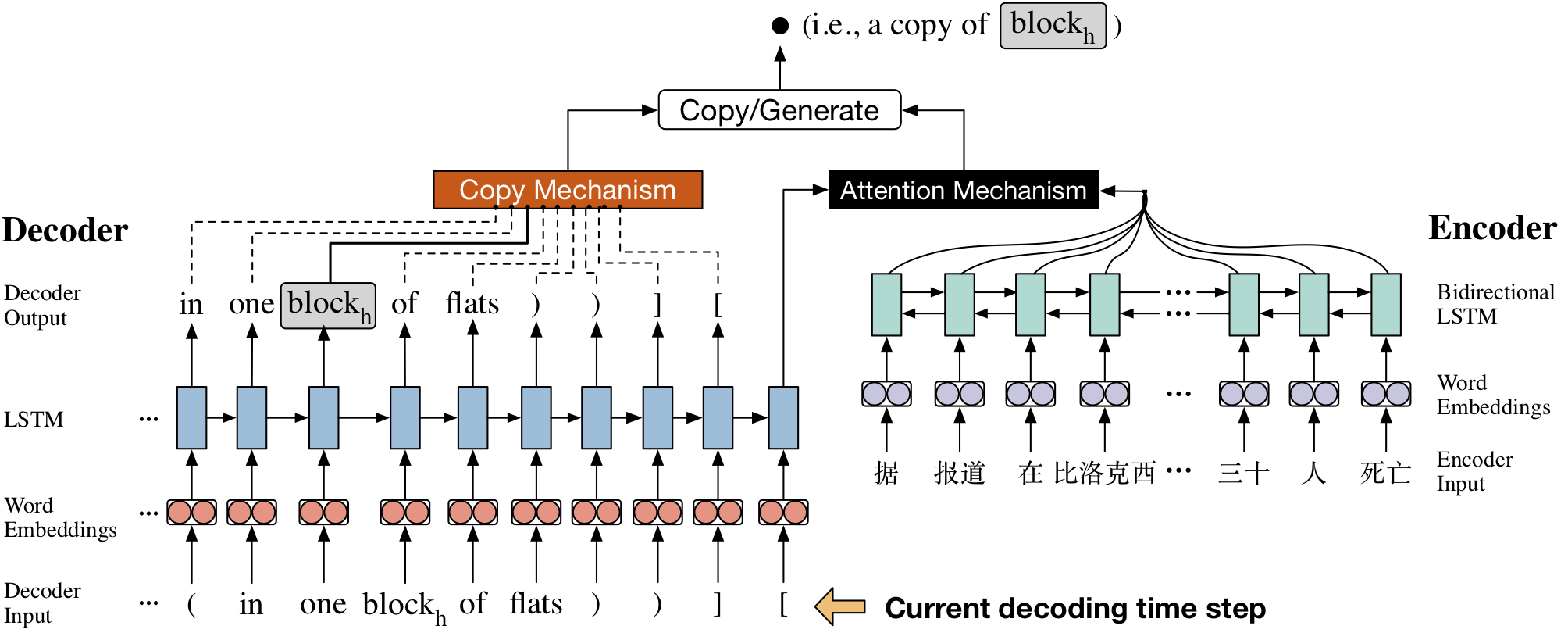}
\caption{Illustration of the model architecture. At the current decoding step,
    the decoder takes a token ``\textbf{[}'' as input, and decides to copy a preceding
    head token ``\textbf{block}$_\textbf{h}$'' via the coping mechanism, instead
    of generating a token via the attention mechanism.
    \label{fig:arch}}
\end{figure*}

The goal for cross-lingual semantic parsing is to learn a conditional
probability distribution $P(Y,A|X)$ whose
most likely configuration, given the input sentence, outputs the true
linearized meaning representation.
While the standard encoder-decoder framework shows the state-of-the-art 
performance in sequence-to-sequence learning~\cite{vinyals2015grammar,jia-liang:2016:P16-1,barzdins-gosko:2016:SemEval},
it can not directly solve intra-sentential conference in cross-lingual semantic parsing.
To achieve this goal, we propose an encoder-decoder architecture incorporated with a {\em copying mechanism}.
As illustrated in~\Cref{fig:arch}, \textbf{Encoder} transforms the input sequence
into hidden states; \textbf{Decoder} reads the hidden states, and 
then at each time step decides whether to generate a token or
copy a preceding token.

\subsection{Encoder}
The encoder employs a bidirectional recurrent neural
network~\cite{schuster1997bidirectional} to encode the input
$X=x_1,\ldots,x_{N}$\footnote{For simplicity,
we use $X$ (and $Y$) to represent both tokens as well as their word embeddings.}
into a sequence of hidden states
$\bm{h} = h_1,\ldots,h_{N}$.
Each hidden state $h_i$ is a concatenation of a left-to-right
hidden state $\overrightarrow{h_i}$ and a right-to-left hidden state
$\overleftarrow{h_i}$,
\begin{align}
    h_i = \left [
        \begin{aligned}
        \overleftarrow{h}_i\\
        \overrightarrow{h}_i
    \end{aligned}
\right ] = \left [
    \begin{aligned}
        \overleftarrow{f}(x_i, \overleftarrow{h}_{i+1})\\
        \overrightarrow{f}(x_i, \overrightarrow{h}_{i-1})
    \end{aligned}
    \right], \label{eq:encoder}
\end{align}

where $\overleftarrow{f}$ and $\overrightarrow{f}$ are $L$-layer stacked
LSTM units~\cite{hochreiter1997long}. 
The encoder hidden states are zero-initialized.

\subsection{Copying-Enabled Decoder}
Given the encoder hidden states, the decoder predicts meaning representation
according to the conditional probability $P(Y,A\mid X)$ which can be
decomposed as a product of the decoding probabilities at each time step $t$:
\begin{equation}
    P(Y, A \mid X) = \prod\limits_{t=1}^{M}P(y_t, a_t \mid y_{<t}, a_{<t}, X)
\end{equation}
where $y_{<t}$ and $a_{<t}$ are the preceding tokens and the coreference assignments.
We omit $y_{<t}$ and $a_{<t}$ from the notation when the context is unambiguous.
The decoding probability at each time step $t$ is defined as
\begin{equation}
  P(y_t, a_t)=\begin{cases}
      P_{\mathrm{g}}(y_t), & \text{if $a_t=\epsilon$}\\
      P_{\mathrm{c}}(y_t), & \text{otherwise}
  \end{cases}
\end{equation}
where $P_{\mathrm{g}}$ is the generating probability, and $P_{\mathrm{c}}$ is the
copying probability. If the dummy antecedent $\epsilon$ is assigned to $y_t$,
the decoder generates a token for $y_t$, otherwise the decoder copies a token
from the preceding tokens.

\noindent{\bf Generation:} If the decoder decides to generate a token at time
step $t$, the probability distribution of the generated token $y_t$ is defined as
\begin{equation}
    P_{\mathrm{g}}(y_t) = \text{softmax}(\textsc{ffnn}_{\mathrm{g}}(s_t, c_t))
\end{equation}
where $\textsc{ffnn}_{\mathrm{g}}$ is a two-layer feed-foward neural network
over the decoder hidden state $s_t$
and the context vector $c_t$.
The decoder hidden state $s_t$ is computed by
\begin{equation}
    s_t = \textsc{rnn}(y_{t-1}, s_{t-1}) \label{eq:decoder-rnn}
\end{equation}
where \textsc{rnn} is a recurrent neural network using $L$-layer stacked LSTM,
and $s_0$ is initialized by
the last encoder left-to-right hidden state $\overrightarrow{h}_N$.
The context vector $c_t$ is computed by \textbf{Attention Mechanism}
\cite{bahdanau2014neural,luong-pham-manning:2015:EMNLP} as illustrated in
\Cref{fig:arch},
\begin{align}
    c_{t} & = \sum_{i}^{N}\alpha_{t,i}h_i, \\
    \alpha_{t,i} & = \frac{\exp{(s_{t}^\top (W_{\alpha}h_i + b_\alpha)))}}{\sum_{j=1}^{N}\exp{(s_{t}^\top (W_{\alpha}h_{j} + b_\alpha))}},
\end{align}
where $W_\alpha$ is a transform matrix and $b_\alpha$ is a bias.

\noindent{\bf Copying Mechanism:} If the decoder at time step $t$ decides to copy a token
from the preceding tokens as shown in \Cref{fig:arch},
the probability of $y_t$ being a copy of the preceding token $y_k$ is defined as
\begin{align}
    P_{\mathrm{c}}(y_t=y_k) = \frac{\exp{(\textsc{score}(y_{t}, y_k))}}{\sum_{y_k' \in \mathcal{A}(t)}\exp{(\textsc{score}(y_{t}, y_{k'}))}},
\end{align}
where $\mathcal{A}(t)=\{\epsilon, y_1,\ldots,y_{t-1}\}$ is the set of possible
coreference assignments for $y_t$ defined in \Cref{sec:task}. $\textsc{score}(y_t, y_k)$ is a pairwise
score for a coreference link between $y_t$ and $y_k$. There are three terms in
this pairwise coreference score, which is akin to~\newcite{lee-EtAl:2017:EMNLP2017}:
(1) whether $y_t$ should be a copy of a preceding token,
(2) whether $y_k$ shoud be a candidate source of such a copy,
and (3) whether $y_k$ is an antecedent of $y_t$.
\begin{align}
    \textsc{score}(y_t, y_k) = s_{\mathrm{c}}(y_t) + s_{\mathrm{p}}(y_k) + s_{\mathrm{a}}(y_t, y_k) 
\end{align}
Here $s_{\mathrm{c}}(y_t)$ is a unary score for $y_t$ being a copy of a preceding
token, $s_{\mathrm{p}}(y_k)$ is a unary score for $y_k$ being a candidate source of
such a copy, and $s_{\mathrm{a}}(y_t, y_k)$ is a pairwise score for 
$y_k$ being an antecedent of $y_t$. 

\begin{figure}[t]
\centering
\includegraphics[width=0.4\textwidth]{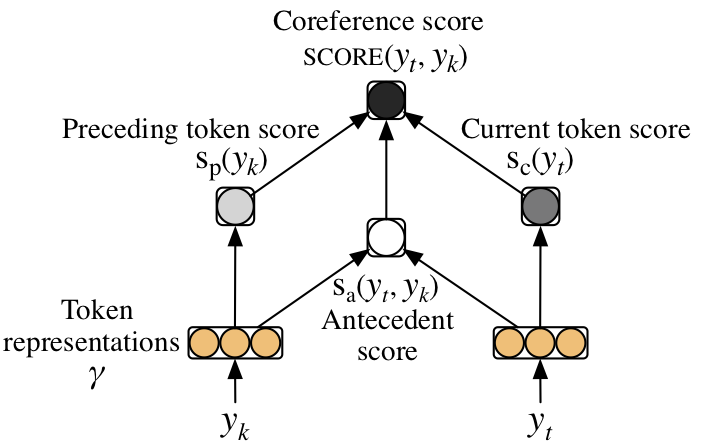}
\caption{Coreference scoring architecture in the copy mechanism 
between a preceding token $y_k$ and the currently considered token $y_t$.
    \label{fig:copy}}
\end{figure}

\Cref{fig:copy} shows the details of the scoring architecture in the copy
mechanism. At the core of the three factors are vector representations
$\gamma(y_t)$ for each token $y_t$, which is described
in detail in the following section. Given the currently considered
token $y_t$ and a preceding token $y_k$, the scoring functions above are computed
via standard feed-foward neural networks:
\begin{align}
    s_{\mathrm{c}}(y_t)  = & w_{\mathrm{c}}\cdot  \textsc{ffnn}_{\mathrm{c}}(\gamma(y_t))  \\
    s_{\mathrm{p}}(y_k)  = & w_{\mathrm{p}}\cdot  \textsc{ffnn}_{\mathrm{p}}(\gamma(y_k))  \\
    s_{\mathrm{a}}(y_t, y_k)  = & w_{\mathrm{a}}\cdot  \textsc{ffnn}_{\mathrm{a}}\big([\gamma(y_t), \gamma(y_k\big),\nonumber \\
    &  \gamma(y_t) \circ \gamma(y_k)]) 
\end{align}
where $\cdot$ denotes dot product, $\circ$ denotes element-wise multiplication,
and \textsc{ffnn} denotes a two-layer feed-foward neural network over the input.
The input of $\textsc{ffnn}_{\mathrm{a}}$ is a concatenation of vector representations
$\gamma(y_t)$ and $\gamma(y_k)$, and their explicit element-wise similarity
$\gamma(y_t) \circ \gamma(y_k)$.

\noindent\textbf{Token representations}: To accurately predict conference scores,
we consider three types of information in each token representation $\gamma(y_t)$:
(1) the token itself $y_t$,
(2) on the decoder side, the preceding context $y_{<t}$,
and (3) on the encoder side, the input sequence $X=x_1,\ldots,x_{N}$.

The lexical information of the token itself $y_t$ is represented by
its word embedding $e(y_t)$.
The preceding context $y_{<t}$ is encoded by the decoder \textsc{rnn} in
\Cref{eq:decoder-rnn}. We use the decoder hidden state $s_t$ to represent
the preceding context information.

The encode-side context is also crucial to predicting coreference:
if $y_t$ and $y_k$ pay attention to the same context on the encoder side,
they are likely to refer the same entity. Therefore, we use the context vector
$o_t$ computed by an attention mechanism to represent the encoder side
context information for $y_t$:
\begin{align}
    o_{t} & = \sum_{i}^{N}\beta_{t,i}h_i, \\
    \beta_{t,i} & = \frac{\exp{(s_{t}^\top W_{\beta}h_i))}}{\sum_{j=1}^{N}\exp{(s_{t}^\top W_{\beta}h_{j})}},
\end{align}
where $h_i$ is the encoder hidden state computed by \Cref{eq:encoder},
$s_t$ is the decoder hidden state computed by \Cref{eq:decoder-rnn}, and
$W_{\beta}$ is a transform matrix.

All the above information is concatenated to produce the final token 
representation $\gamma(y_t)$:
\begin{align}
    \gamma(y_t) = [e(y_t), o_t]
\end{align}

\subsection{Learning}
In the training objective, we consider both the copying accuracy as well as
the generating accuracy. Given the input sentence $X$, the output sequence
of tokens $Y$, and  the coreference assignments $A$,
the objective is to minimize the negative log-likelihood:
\begin{align}
    \mathcal{L} = & -\frac{1}{|\mathcal{D}|}\sum\limits_{(X, Y, A)\in\mathcal{D}}\log P(Y, A \mid X) \nonumber \\
    = & -\frac{1}{|\mathcal{D}|}\sum\limits_{(X, Y, A)\in\mathcal{D}}\sum\limits_{t=1}^M[P_{\mathrm{g}}(y_t) + \mu P_{\mathrm{c}}(y_t)] \nonumber 
\end{align}

To increase the convergence rate, we pretrain the model by setting $\mu=0$
to only optimize the generating accuracy. After the model converges, we set
$\mu$ back to $1$ and continue training. Since most tokens in the output
are not copied from their preceding tokens, and are therefore assigned
the dummy antecedent $\epsilon$, the training of the copy mechanism
is heavily unbalanced.
To alleviate the balance problem, we consider
coreference assignments of syntactic head tokens in optimization.

\section{Experiments}
\label{sec:exp}
We now describe the evaluation data, baselines, and experimental results.
Hyperparameter settings are reported in the supplemental material.

\noindent\textbf{Data}: We choose Chinese as the source language
and English as the target language.
For \textbf{testing}, we sampled 2,258 sentences from Universal Dependencies (UD) English
Treebank~\cite{silveira14gold}, which is taken from five genres of web media:
weblogs, newsgroups, emails, reviews, and Yahoo answers.
We then created PredPatt meaning representations for these sentences based
on the gold UD annotations. Meanwhile, the Chinese translations of these sentences
were created by crowdworkers on Amazon Mechanical Turk.
The test dataset will be released upon publication.
For \textbf{training}, we first collected about 1.8M Chinese-English sentence
bitexts from the GALE project~\cite{cohen2007gale},
then tokenized Chinese sentences with Stanford Word Segmenter~\cite{chang-galley-manning:2008:WMT}.
We created PredPatt meaning representations for English sentences based
on automatic UD annotations generated by SyntaxNet Parser~\cite{andor-EtAl:2016:P16-1}.
We hold out 10K training sentences for \textbf{validation}.
The dataset statistics are reported in \Cref{tab:data}.

\begin{table}[h]
\centering
\small
\begin{tabular}{@{}lcc@{}}
\toprule
           & No. sentences      & Source      \\ \midrule
Train      & 1,889,172 & GALE        \\
Validation & 10,000    & GALE        \\
Test       & 2,258     & UD Treebank \\ \bottomrule
\end{tabular}
\caption{Statistics of the evaluation data.}
\label{tab:data}
\end{table}

\begin{table*}[htp]
\setlength{\tabcolsep}{2pt}
\centering
\small
\begin{tabular}{@{}l@{\hskip 16pt}ccc@{\hskip 16pt}ccc@{\hskip 16pt}ccc@{\hskip 14pt}c@{}}
\toprule
 & \multicolumn{3}{c}{MUC} & \multicolumn{3}{c}{B$^3$} & \multicolumn{3}{c}{CEAF$_e$} & \multirow{2}{*}{Avg. F$_1$ } \\
 & Prec. & Rec. & F$_1$ & Prec. & Rec. & F$_1$  & Prec. & Rec. & F$_1$  &  \\ \midrule
    \textsc{Seq2Seq+Copy} & \textbf{95.67} & \textbf{96.43} & \textbf{96.05} & \textbf{98.83} & \textbf{99.14} & \textbf{98.99} & \textbf{98.54} & \textbf{98.32} & \textbf{98.43} & \textbf{97.82} \\
    \textsc{Seq2Seq+Heuristics} & 85.67 & 51.45 & 64.29 & 97.94 & 88.23 & 92.83 & 84.24 & 93.68 & 88.71 & 81.94 \\
    \textsc{Seq2Seq+Random} & 10.44 & 31.13 & 15.63 & 37.91 & 83.89 & 52.22 & 62.37 & 27.69 & 38.36 & 35.40 \\ \bottomrule
\end{tabular}
\caption{Evaluation of coreference results on the test dataset. We force the model to decode the gold target sequence,
and only evaluate algorithms for solving coreference. The Avg. F$_1$ is computed by averaging F$_1$ scores of
MUC, B$^3$ and CEAF$_e$.}
\label{tab:coref}
\end{table*}

\begin{table}[t]
\centering
\small
\begin{tabular}{@{}lccc@{}}
\toprule
 & Prec. & Rec. & F$_1$ \\ \midrule
    \textsc{Seq2Seq+Copy} & \textbf{49.72} & \textbf{31.25} & \textbf{38.38} \\
    \textsc{Seq2Seq+Heuristic} & 46.76 & 30.88 & 37.20 \\
    \textsc{Seq2Seq+Random} & 42.86 & 30.74 & 35.80 \\
    \textsc{Pipeline} & 28.50 & 20.65 & 23.95 \\ \bottomrule
\end{tabular}
\caption{Precision, recall, and $F_1$ scores for the evaluation metric $\bm{S}_{(\phi,\psi)}$ on the test dataset.}
\label{tab:smatch}
\end{table}

\noindent\textbf{Comparisons}: We evaluate 4 approaches in the experiments:
(1) \textsc{Seq2Seq+Copy} is our proposed approach, described in \Cref{sec:model}.
(2) \textsc{Seq2Seq+Heuristic} preprocesses the data by replacing the special symbol ``$\bullet$''
with the syntactic head of its antecedent.
During training and testing, it replaces the copying mechanism with a heuristic
that solves coreference by randomly choosing an antecedent among preceding arguments
which have the same syntactic head.
(3) \textsc{Seq2Seq+Random} replaces the copying mechanism by randomly choosing
an antecedent among all preceding arguments.
(4) We also include a \textsc{Pipeline} approach where Chinese sentences are
first translated into English by a neural machine translation system~\cite{2017opennmt}
and are then annotated by a UD parser~\cite{andor-EtAl:2016:P16-1}.
The final meaning representations of \textsc{Pipeline} are created based the automatic UD annotations.

\noindent\textbf{Results}:
\Cref{tab:smatch} reports the test set results based on the evaluation metric 
$\bm{S}_{(\phi,\psi)}$ defined in \Cref{sec:metric}.
Overall, our proposed approach, \textsc{Seq2Seq+Copy}, achieves the best precision,
recall, and $F_1$.
The two baselines, \textsc{Seq2Seq+Heuristic} and \textsc{Seq2Seq+Random}
also achieve reasonable results. These two baselines both employ sequence-to-sequence
models to predict meaning representations, which can be considered a replica 
of state-of-the-art approaches for structured prediction
\cite{vinyals2015grammar,barzdins-gosko:2016:SemEval,peng-EtAl:2017:EACLlong1}.
Our proposed approach outperforms these two strong baselines:
the copying mechanism, aiming for solving coreference in cross-lingual semantic
parsing, results in both precision and recall gains.
The detailed gains achieved by the copying mechanism are discussed in the following section.
In the \textsc{Pipeline} approach, each component is trained independently.
During testing, residual errors from each component are propagated through the pipeline.
As expected, \textsc{Pipeline} shows
a significant performance drop compared to the other end-to-end learning approaches.

Coreference occurs 5,755 times in the test data. To evaluate
the coreference accuracy of these end-to-end learning approaches,\footnote{\textsc{Pipeline}
predicts PredPatt meaning representations based on automatic UD annotation,
so its conference accuracy can not be separately evaluated.}
we force each approach to generate the gold target sequence, and only predict
coreference via the \textsc{Copying} mechanism, the \textsc{Heuristic} baseline, or the \textsc{Random} baseline. 
We report the precision, recall, and $F_1$ for the standard MUC, B$^3$, and CEAF$_e$
metrics using the official coreference scorer
of the CoNLL-2011/2012 shared tasks
\cite{pradhan-EtAl:2014:P14-2}.

\Cref{tab:coref} shows the evaluation results.
Since coreference in our setup occurs at the sentence level, the proposed
copying mechanism achieves high scores in all three metrics, and the average
$F_1$ is 97.82.
The heuristic baseline, which solves coreference only based on syntactic
heads, also achieves a relatively high average $F_1$ of 81.94.
Under the MUC metric, the copying mechanism performs significantly 
better than the heuristic baseline.
The random baseline limits the choice of coreference
to preceding syntactic heads, ignoring all other tokens,
achieving scores much lower than the other two approaches in all three
metrics.

\section{Conclusions}
We introduce the task of cross-lingual semantic parsing, which maps content provided in a source language into a meaning representation based on a target language.  
We present: the PredPatt meaning representation as the target semantic interface,
the $\bm{S}_{(\phi,\psi)}$ metric for evaluation,
and the Chinese-English semantic parsing dataset.
We propose an end-to-end learning approach with a copying mechanism which 
outperforms two strong baselines in this  task.
The PredPatt meaning representation, the evaluation metric $\bm{S}_{(\phi,\psi)}$, and the evaluation dataset
provided in this work will be beneficial to the increasing interests in meaning representations and cross-lingual
applications.

\bibliography{acl2018}
\bibliographystyle{acl_natbib}

\end{document}